\documentclass{article}
\usepackage{spconf,amsmath,graphicx}
\usepackage[utf8]{inputenc}
\usepackage[english]{babel}
\usepackage{amssymb}
\usepackage{multirow}
\usepackage{bm}
\usepackage{bbm}
\usepackage{svg}
\usepackage{booktabs}
\usepackage[titletoc,title]{appendix}
\usepackage[dotinlabels]{titletoc}
\usepackage[style=ieee, citestyle=numeric-comp, backend=biber]{biblatex}
\usepackage[font=small,labelfont=bf, margin=1cm]{caption}
\newcommand{\matr}[1]{\underline{\mathbf{#1}}}
\newcommand{\vect}[1]{\mathbf{#1}}
\usepackage{mathtools,etoolbox}

\title{Out-of-Distribution Detection of Melanoma using Normalizing Flows}
\name{M.M.A. Valiuddin, C.G.A Viviers}
\address{Department of Electrical Engineering, Eindhoven University of Technology}
\begin{document}
%
\maketitle

\begin{abstract}
\textbf{Generative modelling has been a topic at the forefront of machine learning research for a substantial amount of time. With the recent success in the field of machine learning, especially in deep learning, there has been an increased interest in explainable and interpretable machine learning. The ability to model distributions and provide insight in the density estimation and exact data likelihood is an example of such a feature. Normalizing Flows (NFs), a relatively new research field of generative modelling, has received substantial attention since it is able to do exactly this at a relatively low cost whilst enabling competitive generative results. While the generative abilities of NFs are typically explored, we focus on exploring the data distribution modelling for Out-of-Distribution (OOD) detection. Using one of the state-of-the-art NF models, GLOW, we attempt to detect OOD examples in the ISIC dataset. We notice that this model under performs in conform related research. To improve the OOD detection, we explore the masking methods to inhibit co-adaptation of the coupling layers however find no substantial improvement. Furthermore, we utilize Wavelet Flow which uses wavelets that can filter particular frequency components, thus simplifying the modeling process to data-driven conditional wavelet coefficients instead of complete images. This enables us to efficiently model larger resolution images in the hopes that it would capture more relevant features for OOD. The paper that introduced Wavelet Flow mainly focuses on its ability of sampling high resolution images and did not treat OOD detection. We present the results and propose several ideas for improvement such as controlling frequency components, using different wavelets and using other state-of-the-art NF architectures.}
\end{abstract}


\section{Introduction}
\label{sec:intro}
Research in probabilistic generative modeling is receiving significant attention. This, because rapid advances in computational power in combination with an increased availability of data makes modeling complex data distributions more feasible. Recently developed techniques show the feasibility of modelling the probability distribution $p_X$ from observed data $\{\vect{x}\}^N_{i=1}$. The applications of the generative models include density estimation, outlier detection, sampling new data and much more. Models such as Variational Auto-encoders (VAEs)~\cite{vae} and Auto-Regressive Generative Neural Networks (such as pixelRNN~\cite{pixelrnn}) have shown impressive attempts at learning the data distribution. Each of these implementations have shown a unique generative ability, but they also have their own set of drawbacks. For instance, in the case of VAEs, we approximate the density of the data through minimizing the evidence lower bound (ELBO). This design choice will rarely make the estimated distribution closely match the actual data distribution and sacrifices tractability during sampling and inference. On the other hand, Generative Adversarial Networks (GANs) do not model the density of the data explicitly but rather focus on the ability to generate realistic images through a mini-max game between a discriminator a generative network. The discriminator can then try to distinguish between in and out of distribution data with no tractable density. This makes qualitative evaluation of the (implicit) distribution not possible. Another downside that comes with it is difficulty during training. Mode collapse, posterior collapse, vanishing gradients and training instability~\cite{gan2,gan3} make them a less reliable solution. Auto-Regressive Generative models in fact, do explicitly model the density. The generated samples are relatively good yet the process very slow due to its sequential data generation.
    
In the research community there has been a trend towards explainable low cost models with tractable probabilities. Normalizing Flows (NF) are capable of sampling as well as do density estimation with tractable probabilities and is gaining significant interest. An impressive characteristic of this architecture is the fact that NFs attempt to model any complex probability distribution~\cite{nf1,nf2,nf3} through likelihood maximization. Despite the NFs being a fairly recent research area, it has produced competitive results on public benchmark datasets. With the recent developments in the NFs architectures~\cite{glow, chen2019neural, waveletflow}, the qualitative and quantitative results as well as computational requirements are extremely competitive. The applications include but are not limited to image generation~\cite{flow++, glow}, noise modelling ~\cite{noiseflow}, video generation~\cite{videoflow}, audio generation~\cite{audioflow, audioflow2, audioflow3}, graph generation ~\cite{graphnvp}, reinforcement learning~\cite{reinflow, reinflow2, reinflow3}, computer graphics~\cite{neuralimportantcesampling} and physics~\cite{physics, physics2, physics3, physics4, physics5}. 

In this research project the Out-of-Distribution (OOD) detection using state-of-the-art models GLOW and Wavelet Flow is investigated. This is done on public datasets as well as The International Skin Imaging Collaboration (ISIC) dataset~\cite{isic}. The dataset contains images of benign as well as malignant melanoma. To the best of our knowledge, the ability of OOD detection with NFs on medical imaging has not been investigated. Since in most practical scenarios anomaly data is scarce, we choose to model the benign images and consider the malignant images anomalous. We find that OOD detection using the GLOW architecture performs poorly due to the exploitation of local pixel correlations and the co-adaptation of the coupling layers. Supported by recent research by Kirichenko \textit{et al.}~\cite{whytho}, the cycle masking technique in the coupling layers are tested an an attempt to inhibit co-adaptation of coupling layers but provide limited improvement. 

Recently, a model named Wavelet Flow~\cite{waveletflow} has been published. Where instead of the data distribution, the detail coefficients are modeled from the low frequency Haar wavelet component. It is shown that wavelets in combination NFs provide numerous benefits including likelihood estimation and sampling at a higher resolutions. The authors showcase this by sampling at a 8x times higher resolution than that of the training images. Furthermore, the proposed model delivers competitive results on public benchmark datasets in bits per dimension with up to 15x faster training. We see wavelets as an opportunity to control the components of the images being modeled by the NFs and leverage it for OOD detection at higher resolution. OOD detection using Wavelet Flow has not been investigated, even the original published work. Thus, we decided to investigate this by applying it to the ISIC dataset and report our findings.

\section{Related work}
Kirichenko \textit{et al.}~\cite{whytho} investigate OOD detection and \textit{Why Normalizing Flows Fail to Detect
Out-of-Distribution Data}. The paper argues that NFs mainly learn local pixel correlations and generic image-to-latent-space transformations instead of capturing semantic information of the images. Furthermore, the authors argue that coupling NFs suffer from co-adaptation, where it can equally likely predict OOD data due to dependency of the coupling layers hence assigning high probabilities to OOD data. One of the proposed solutions cycle masking, has proven to tackle this issue. However, the effectiveness of this method shows to be dependent on the datasets being used. We implement cycle masking in our research in order to determine if it can improve OOD detection for the ISIC dataset.

Ren \textit{et al.}~\cite{ood-likelihood} observe that the likelihood score in OOD detection from deep generative models are often heavily affected by population level background statistics. They propose a likelihood ratio method to effectively correct this. We draw our inspiration to use wavelets to control what components of the image NFs learn. 

\section{Methodology} 
This internship research aims to answer the following questions:
\begin{itemize}
    \item Can we reproduce some of the research of Out-of-Distribution detection with Normalizing Flows and apply it to the ISIC melanoma dataset?
    \item Which techniques can we employ to improve Out-of-Distribution detection on the ISIC melanoma dataset?
    \item Does Wavelet Flow outperform GLOW in Out-of-Distribution detection on the ISIC melanoma dataset?
\end{itemize}

In order to answer these questions, we investigated the theory behind NFs. We apply our newly acquired knowledge of Normalizing Flows to Out-of-Distribution on the ISIC dataset.  We trained the chosen models on the benign images of the dataset and evaluate it against the malignant classes.

\section{Normalizing Flows}
\label{sec:nf}
\begin{figure}[htb]
\begin{minipage}[b]{1.0\linewidth}
  \centering
  \centerline{\includegraphics[width=8.5cm]{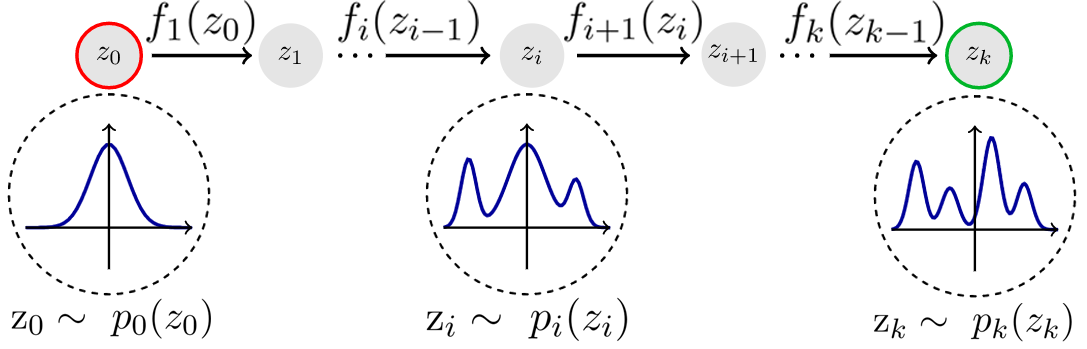}}
  \caption{Series of consecutive transformations transforming a Gaussian base to a complex non-Gaussian target distribution. Image taken from \textit{Janosh Dev.}~\cite{flowimg}}
  \label{fig:flow}
\end{minipage}
\end{figure}

We call a sequence of bijective transformations a Normalizing Flow. NFs map a complex distribution $p_{\vect{x}}(\mathrm{X})$ to a more simple distribution $p_{\vect{z}}(\mathrm{Z})$ (normalizing direction) through a chain of transformations. By constraining these transformations to be bijective, an inverse transformations exist that transforms the base distribution $p_0(\vect{z}_0)$ to the complex distribution $p_k(\vect{z}_k)$ (generative direction) and thus enabling sample generation.  Since we want to map $p_{\vect{x}}(X)$ from $p_{\vect{z}}(Z)$ to enable data generation, it is required that these transformations are invertible.  This idea is visualized in Figure \ref{fig:flow}, where $k$ equals the final complex transformed distribution while $i$ denotes the $i$th intermediate distribution of the chain. In Section \ref{sec:mathback} a mathematical formulation on NFs are provided. This is followed by an overview of different types of flows in  Section \ref{sec:architectures}. The NF architectures implemented during this internship project are presented in Section \ref{sec:glow} and Section \ref{sec:waveletflow}. 
\subsection{Mathematical background}
\label{sec:mathback}
A bijection between distributions should adhere to the law of total probability as 
    \begin{equation}
        \int p(x)dx= \int p(y)dy.    
    \end{equation}
We are only concerned in the absolute change of volume, hence the sign of it is disregarded. We can represent a sample in $Y$ in terms of a sample in $X$. As shown below in the change of variables formula (Eqn.~\ref{change of var}), we can evaluate $p_{\vect{y}}(Y)$ and maintain the total density through any transformation.   
    \begin{equation}
        p(y)=p(x)\left|\frac{dx}{dy}\right|.
        \label{change of var}
    \end{equation}
This local-linear change approximation is in fact the Jacobian determinant. Given that $y=f(x)$ and thus $x=f^{-1}(y)$ we can write
\begin{equation}
    p_Y(y)=p_X(f^{-1}(y))\left|\operatorname{det}\frac{\partial f^{-1}}{\partial y}\right|.
\end{equation}
We denote a sample and its transformation as
\begin{equation}
    \begin{aligned}
    \mathbf{z}_{i-1} & \sim p_{i-1}\left(\mathbf{z}_{i-1}\right) \\
    \mathbf{z}_{i} &=f_{i}\left(\mathbf{z}_{i-1}\right), \mathbf{z}_{i-1}=f_{i}^{-1}\left(\mathbf{z}_{i}\right). \\
    \end{aligned}
\end{equation}
Given the chain of the transformations
\begin{equation}
\mathbf{x}=\mathbf{z}_{K}=f_{K} \circ f_{K-1} \circ \cdots f_{1}\left(\mathbf{z}_{0}\right)
\end{equation}
the log-likelihood of the observed data $\vect{x}$ is concisely written as
\begin{equation}
    \begin{aligned}
        \log p(\mathbf{x})=\log p_{0}\left(\mathbf{z}_{0}\right)-\sum_{i=1}^{K} \log \left|\operatorname{det} \frac{d f_{i}}{d \mathbf{z}_{i-1}}\right|
    \end{aligned}
\end{equation}
See appendix \ref{A} for the complete derivation. Notice how the log-likelihood of $p(\vect{x})$ only requires inference on the base distribution from the latent vector $\vect{z_0}$. All other contributions to the log-likelihood are from the absolute value of Jacobian determinant of the bijections. An ideal bijection in NFs should be:
\begin{itemize}
\item invertible in all cases
\item cheap to invert
\item have an easy to compute Jacobian determinant
\item expressive enough to acquire the target distribution.
\end{itemize}

For evaluation, bits per dimension (BPD) is used. BPD is defined as
\begin{equation}
    bpd(\vect{x}) = \frac{\operatorname{log} p(\vect{x})}{W \times H \times C \times \operatorname{log}(2)}
\end{equation}
 where $H$, $W$ and $C$ stand for height, width and channels of the image.

\subsection{Architectures}
\label{sec:architectures}
Several families of bijections exist with each having their own (dis-) advantages. Kobyzef~\textit{et al.}~\cite{nf1} provides extensive overview of NFs and their performance on the various benchmarks. In the following section we introduce basic flows and their limitations. We define coupling flows in more detail as they significantly improve on their predecessors.
\subsubsection{Simple flows}
Elementwise bijections are the most simple form of transformations in NFs. Let $h: \mathbb{R} \mapsto \mathbb{R}$ be a bijection of scalar value. Then given vector $\vect{x} \in \mathbb{R}$ we define a flow as
    \begin{equation}
        \begin{aligned}
            g(\vect{x})=(h(x_1), h(x_2),...,h(x_D))^T.
        \end{aligned}
    \end{equation}
Here the Jacobian determinant is just the product of the absolute values of the derivative of $h$. A logical extension towards element-wise flows are linear mappings as
\begin{equation}
    g(\vect{x})=\matr{A}\vect{x}+\vect{b},
\end{equation}
where the observation matrix and bias are $\matr{A} \in \mathbb{R}^{DxD}$ and $\mathbf{b} \in \mathbb{R}^{D}$, respectively. Since the Jacobian determinant is $\det(\matr{A})$, several restrictions can be placed on the observation matrix to make the computation of the Jacobian determinant more efficient, such as forcing it to be diagonal or triangular. Furthermore, \textit{LU decomposition} on the observation matrix has shown to reduce computation complexity to $\mathcal{O}(D)$ and $\mathcal{O}(D^2)$ in the forward and backwards pass, respectively \cite{triang}. Element-wise and linear flows are not expressive enough to capture complex distributions. The output of these flows are always in the same family of distributions as the input. This is because the flows are not able to expand and contract densities. Planar flows and radial flows~\cite{nfvae} are able to do this. The bijection of the planar flow can be written as

\begin{equation}
        g(\vect{x})=\vect{x} + \vect{u}h(\vect{w}^T\vect{x}+\vect{b}).
    \end{equation}

For radial flows, the expansion and contraction is around a particular point as

\begin{equation}
        g(\vect{x})=\vect{x} + \frac{\beta}{\alpha \left|\vect{x}-\vect{x}_0\right|}(\vect{x}-\vect{x}_0).
\end{equation}

Both flows need many consecutive bijections in order to model complex distributions. This can get computationally expensive. Also, the invertibility of the flow holds for only specific conditions. Their practicality has mostly shown in variational inference, as is done in the cited publication.
\subsubsection{Coupling Flows}
\label{coupling}

\begin{figure}[tp]
\begin{minipage}[b]{.48\linewidth}
  \centering
  \centerline{\includegraphics[width=4.0cm]{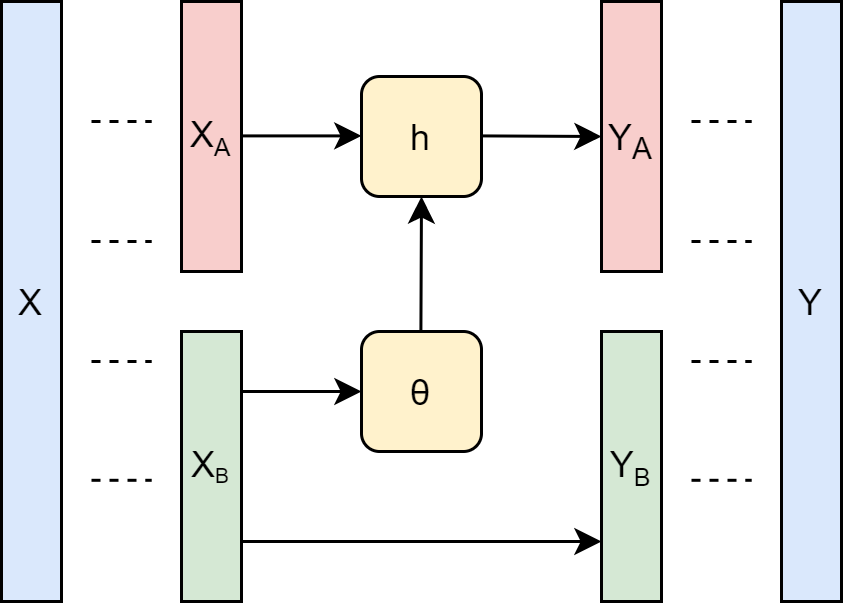}}
  \centerline{\textbf{A}}
\end{minipage}
\hfill
\begin{minipage}[b]{0.48\linewidth}
  \centering
  \centerline{\includegraphics[width=4.0cm]{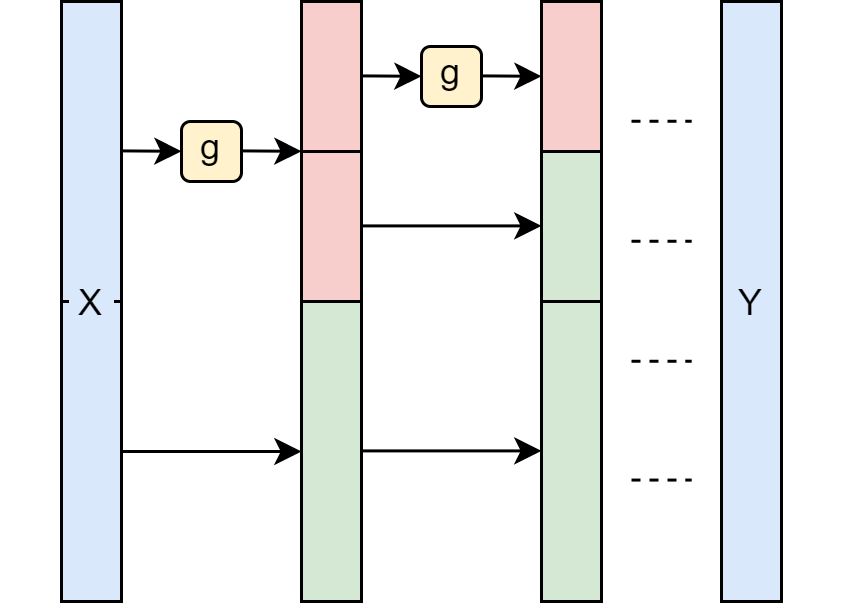}}
  \centerline{\textbf{B}}
\end{minipage}
\caption{(A) Conventional and (B) multi-scale coupling architecture in the normalizing direction}
\label{fig:coupling}
\end{figure}

Dinh \textit{et al.}~\cite{nice} introduced a method for highly expressive transformations referred to as \textit{coupling} flows. Before passing through the bijection, the input $\vect{x} \in \mathbb{R}$ will be split into two subspaces $\vect{x}_A \in \mathbb{R}^d$ and $\vect{x}_B \in \mathbb{R}^{D-d}$. The bijection function is $h(\bullet, \theta(\bullet)): \mathbb{R}^d \mapsto \mathbb{R}^d$. With the \textit{conditioner} $\theta$ being any arbitrary function, we can regard the coupling flow $g:\mathbb{R}^D \mapsto \mathbb{R}^D$ as functions
\begin{equation}
\begin{aligned}
    \vect{y}_A &= h(\vect{x}_A,\theta(\vect{x}_B))\\
    \vect{y}_B &= \vect{x}_B.
\end{aligned}
\end{equation}
For the invertibility of the total flow we require the conditioner to be invertible as well. The inverse can be stated as
\begin{equation}
\begin{aligned}
    \vect{x}_A &= h^{-1}(\vect{y}_A,\theta(\vect{x}_B))\\
    \vect{x}_B &= \vect{y}_B.
\end{aligned}
\end{equation}
If we consider the Jacobian
\begin{equation}
\matr{J}_h=\left[\begin{array}{cc}
\frac{\partial\vect{y}_A}{\partial\vect{x}_A} & \frac{\partial \vect{y}_A}{\partial x_{B}} \\
0 & \mathbbm{I}_d
\end{array}\right]
\end{equation}
it can be seen that the determinant equals $\det \matr{J}_h = \frac{\partial \vect{y}_A}{\partial \vect{x}_B}$. The Jacobian is a triangular matrix in the case of a element-wise operating conditioner. See Figure~\ref{fig:coupling} for an illustration of the conventional coupling flow. Most state-of-the-art models are affine coupling-based NFs such as Glow (section \ref{sec:glow}) and variants of it (Section \ref{sec:waveletflow}). Other methods such as Nonlinear squared, Continuous Mixed CDF, Spline, Neural Autoregressive, Sum-of-Square and Real-and-Discrete coupling flows~\cite{nlsf, flow++, neuralimportantcesampling, cubic, neuralsplinef, neuralautoregressive, sos, rad} exist but has not seen equal success.

In regular coupling flows, the conditioner is always dependent on the split vector  $\vect{x}_B$. It is possible to have the conditioner be independent of $\vect{x}_B$ to develop something called a \textit{multi-scale} flow~\cite{realnvp} (Figure \ref{fig:coupling}). A \textit{multi-scale} flow introduces dimension to a bijective transformation $g$ in the generative direction. In the normalizing direction, the dimension of the vector that undergoes the bijection $g$ decreases. This is analogous to some degree with filters in convolutional neural networks (CNN) when the coupling vector becomes smaller as we traverse deeper in the network where it enables the network to capture more multi-scale information. This works especially well with natural images of landscapes because these images often contain a general structure over the whole canvas.

\subsection{Glow}
\label{sec:glow}
GLOW~\cite{glow} is a coupling NF which has proven to obtain very competitive results compared to other generative models. GLOW builds on improvements introduced by NICE and RealNVP~\cite{realnvp, nice}. RealNVP improved upon the NICE architecture by introducing data dependent shift and translation parameters $s,t: \mathbb{R}^d \mapsto \mathbb{R}^{D-d}$ to use in the coupling layer as
\begin{equation}
\begin{aligned}
    \vect{y}_A &= \vect{x}_A \cdot e^{s(\vect{x}_B)} +t(\vect{x}_B)\\
    \vect{y}_B &= \vect{x}_B
\end{aligned}
\end{equation}
with the inverse then being
\begin{equation}
\begin{aligned}
    \vect{x}_A &= (\vect{y}_A - t(\vect{x}_B)) \cdot e^{-s(\vect{x}_B)}\\
    \vect{x}_B &= \vect{y}_B.
\end{aligned}
\end{equation}
This is contrary to NICE, where instead $\vect{y}_B$ is directly passed through an arbitrary conditioner function and added to its counterpart $\vect{y}_A$.
There are three key concepts that GLOW introduces to improve upon RealNVP. The first concept concerns the reduction of noise variance in the activations. Usually, deep models are trained with small batch sizes. This especially holds for training with high-dimensional datasets due to the available memory in the GPUs. To compensate for this, small minibatches are used. However, this introduces greater noise variation in the batch normalization step. The GLOW paper suggest adding an \textit{actnorm} layer that initializes the model such that the activations have zero mean and unit variance and thus reducing the activation noise variation. After initialization, the parameters will adapt during training like the other parameters of the model. To achieve the goal of normalization, the actnorm layer introduces affine transformations using per-channel scale and bias parameters with function and inverse as
\begin{figure}[htb]
\begin{minipage}[b]{.48\linewidth}
  \centering
  \centerline{\includegraphics[width=3.5cm]{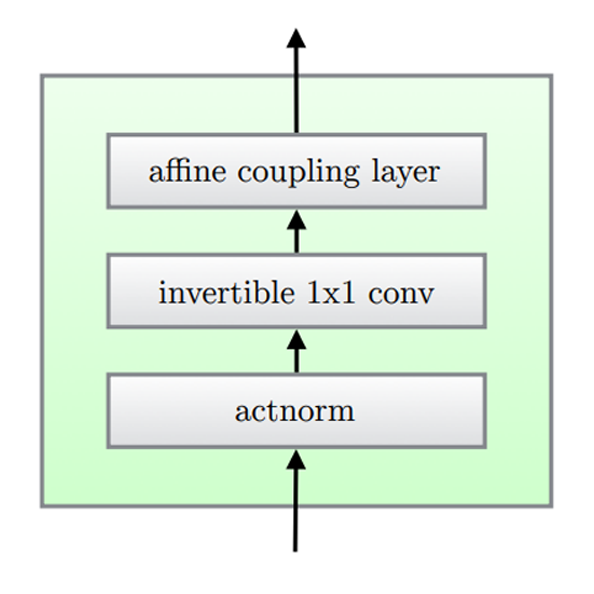}}
  \centerline{\textbf{A} A single flow step}
\end{minipage}
\hfill
\begin{minipage}[b]{0.48\linewidth}
  \centering
  \centerline{\includegraphics[width=3.5cm]{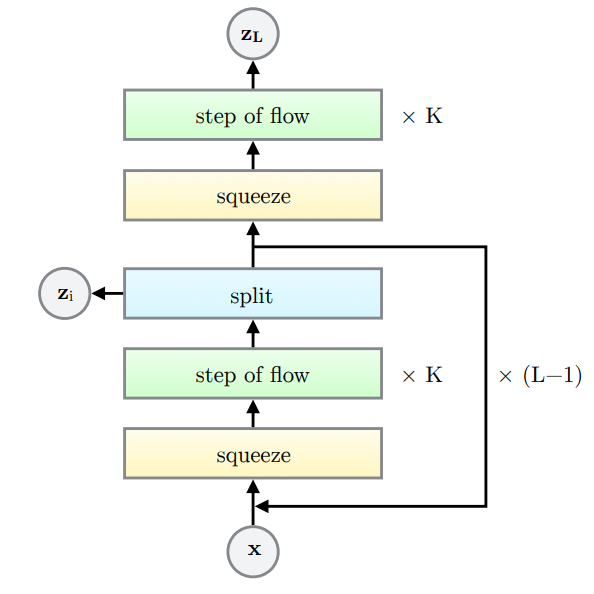}}
  \centerline{\textbf{B} Multi-scale GLOW architecture}
\end{minipage}
\caption{The GLOW architecture as proposed by Kingma \textit{et al.}~\cite{glow}.}
\label{fig:glow}
\end{figure}
\begin{subequations}
\begin{align}
    \vect{y}_{i,j} &= \vect{s} \cdot \vect{x}_{i,j} + \vect{b} \\
    \vect{x}_{i,j} &= \frac{\vect{y}_{i,j} - \vect{b}}{\vect{s}}
\end{align}
\end{subequations}
and log determinant
\begin{equation}
    h \cdot w \cdot \operatorname{sum}(\operatorname{log}\left|\vect{s}\right|).
\end{equation}
Secondly, GLOW suggest using invertible 1x1 convolutional layers with equal input and output channels (which is a generalization of a permutation) instead of reversing the orders of the channels. Now, the permutations can be learned during training. The permutations are done to ensure that every dimension will have the ability to be altered. In combination with \textit{LU decomposition} the computational costs of the log Jacobian determinant can be reduced from $\mathcal{O}(c^3)$ to $\mathcal{O}(c^2)$ as
\begin{equation}
    \operatorname{log}\left|\det(\matr{W})\right| = \operatorname{sum}(\operatorname{log}\left|\vect{s}\right|).
\end{equation}
The final improvement of GLOW is concerning the multi-scale design approach. The GLOW architecture (Figure \ref{fig:glow}) consist of a squeezing function, applying the above mentioned flow for $K$ number of times and then a splitting layer. These three steps are repeated $L$ times and are followed by a final squeeze and flow step. The flow itself first uses the actnorm layer then the invertible 1x1 convolution and finally the affine coupling layer.
The paper compares GLOW against it predecessor and concludes that it outperforms them by great margin on public benchmark datasets such as CIFAR10, ImageNet and LSUN in both image generation and density estimation.

\subsection{Wavelet Flow}
\label{sec:waveletflow}
\begin{figure}[htb]
\begin{minipage}[b]{1.0\linewidth}
  \centering
  \centerline{\includegraphics[width=8.5cm]{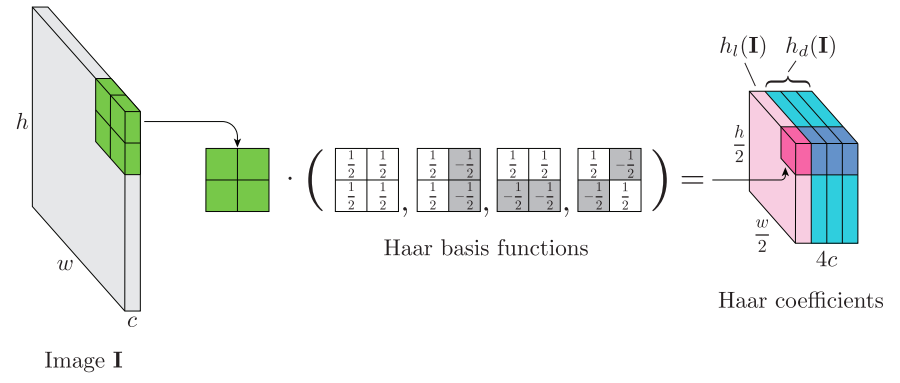}}
  \caption{Example transformation of an image \textbf{I} in the Wavelet Flow architecture with the four Haar basis kernels~\cite{waveletflow}.}
  \label{fig:haartrans}
\end{minipage}
\end{figure}

\begin{figure}[htb]
\begin{minipage}[b]{1.0\linewidth}
  \centering
  \centerline{\includegraphics[width=9cm]{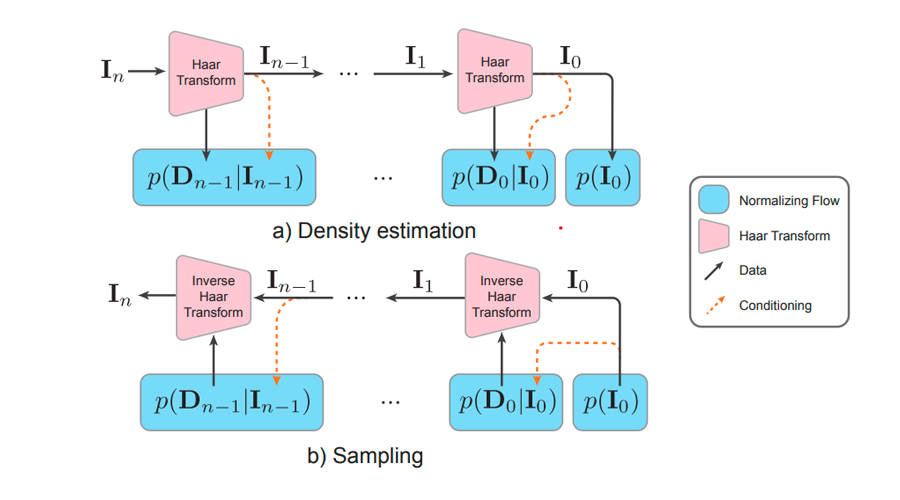}}
  \caption{The Wavelet Flow~\cite{waveletflow} architecture in the normalizing (a) as well as the generative (b) direction. The \textit{i}th level of this model consists of a Haar transform and a Normalizing Flow. The lowest level consists of only a normalizing flow.}
  \label{fig:waveletflow}
\end{minipage}
\end{figure}

In a recent publication by Yu \textit{et al.}~\cite{waveletflow}, they introduce Wavelet Flow, a flow architecture for fast training of high resolution NFs.  They use NFs in combination with Haar wavelets in a multiresolution-like architecture i.e. cascaded discrete wavelet transforms (DWT). In a similar manner, it utilizes these Haar wavelets to downsample and obtains detail coefficients from the input image.
The architecture consists of multiple levels, where at each level the DC component of the Haar basis function is used for the low pass component $h_l$, and the vertical, horizontal and diagonal filters are used to extract high frequency detail coefficients denoted as $h_d$. See Figure \ref{fig:haartrans} for a detailed depiction of the transformation from image \textbf{I} to the Haar coefficients \textbf{D}. The first filter is the DC component i.e. the low pass filter of the wavelet transform.
During training, a NF that closely resembles the GLOW architecture is used to model $h_d$ conditioned on $h_l$. Here, the multi-scale squeeze/split architecture is not used since the wavelets are used to decompose the image at multiple scales. The total multilayer architecture is give in Figure \ref{fig:waveletflow}.  This architecture enables both density estimation and sampling with the additional benefits of:
\begin{itemize}
\item Possibility to do image generation and likelihood estimation at each scale
\item Training and likelihood estimation at each scale \textit{independently}. However, sampling is still sequential.
\item Data driven upsampling of lower resolution images
\item Up to 15x faster training time
\item Competitive with other models in BPD evaluations

\end{itemize}

The summed losses of each individual layer results in the total loss of the Wavelet Flow as
\begin{equation}
    \begin{aligned}
        \log p(\mathbf{I})=\log p_{0}\left(\mathbf{I_0}\right)+\sum_{i=0}^{n-1} \log p(\mathbf{D_i}\mid\mathbf{I_i}).
    \end{aligned}
\end{equation}

\section{Dataset}
\begin{figure}[tp]
\begin{minipage}[b]{.48\linewidth}
  \centering
  \centerline{\includegraphics[width=3.5cm]{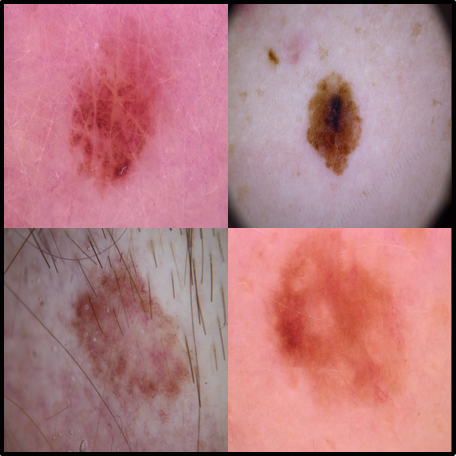}}
  \centerline{\textbf{A} Benign}
\end{minipage}
\hfill
\begin{minipage}[b]{0.48\linewidth}
  \centering
  \centerline{\includegraphics[width=3.5cm]{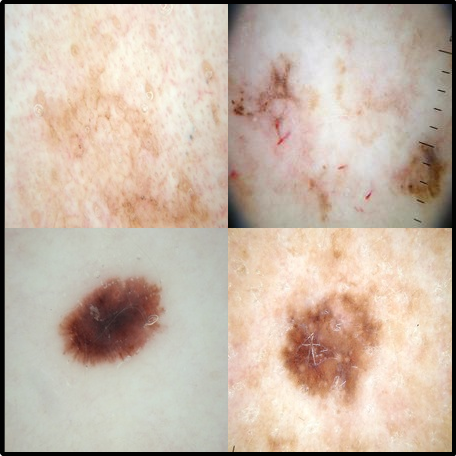}}
  \centerline{\textbf{B} Malignant}
\end{minipage}
\caption{Example images from the adjusted ISIC dataset of melanoma used for OOD detection.}
\label{fig:isic}
\end{figure}

In this research a processed version of the ISIC dataset is used. The dataset consists of 1800 images (224x224) of both benign as well as malignant melanoma. Some example images are depicted in Figure \ref{fig:isic}. The train and test set of the benign images consists of 1440 and 360 images, respectively.

\section{Results}
\label{sec:results}
\begin{figure}[htb]
\begin{minipage}[b]{1.0\linewidth}
  \centering
  \centerline{\includegraphics[width=8.5cm]{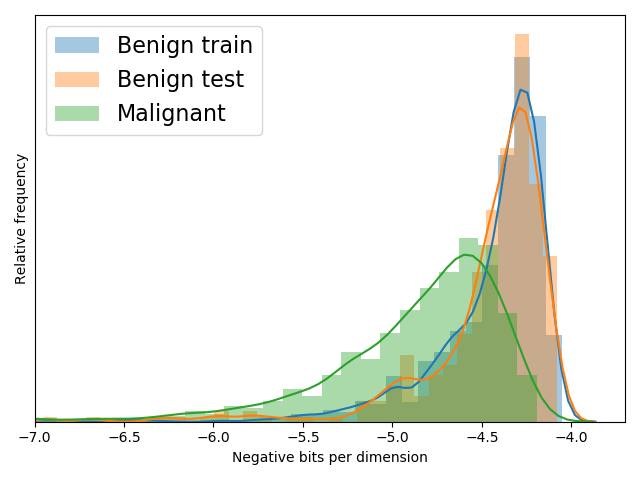}}
  \caption{Likelihood estimation of GLOW masked channel-wise and trained on ISIC benign images.}
  \label{fig:glowres1}
\end{minipage}
\end{figure}

\begin{figure}[htb]
\begin{minipage}[b]{1.0\linewidth}
  \centering
  \centerline{\includegraphics[width=8.5cm]{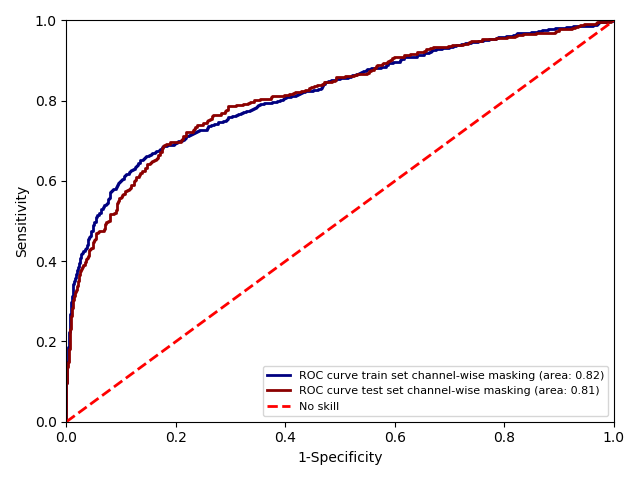}}
  \caption{ROC curves of channel-wise masked GLOW trained on Benign images. The train and test distributions are compared to the malignant out-of-dataset images.}
  \label{fig:ood-roc}
\end{minipage}
\end{figure}

We present our results after experimenting with GLOW and Wavelet Flow. GLOW with parameter settings $L=3$ and $K=32$ is trained on downsampled 64x64 images of the ISIC dataset on a NVIDIA RTX 2080 SUPER. Initially, the model is trained with a batch size of 16, learning rate of 5e-4, weight decay of 5e-5 and 256 hidden channels in the convolutional neural network (CNN) that is responsible for estimating parameters $s$ and $t$. For masking in the coupling layer, we use the channel-wise method. We present our results after 1000 epochs in Figure \ref{fig:glowres1} with its corresponding Receiver Operating Characteristic (ROC) curve in Figure \ref{fig:ood-roc}.
Furthermore, we experimented with channel-wise, checkerboard and cycle masking from Ren et al. [2020]. Cycling is only done for one iteration due to limited dedicated GPU memory. For the shift and translate parameters we use a single hidden layer convolutional neural network (CNN) with ReLU activation except on the output layer. We show the hyperparameter settings in Table \ref{tab:settings}.
\begin{table}[htb]
    \centering
    \begin{tabular}{@{} *4l @{}}    \toprule
     & \textbf{channel-wise} & \textbf{checker} & \textbf{cycle-1} \\ \midrule
     epochs & 1000 & 350 & 500 \\
     batch size    & 16  & 8  & 8 \\ 
     learning rate    & 0.0005  & 0.0001  & 0.0001 \\ 
     weight decay    & 0.001  & 0.001  & 0.001 \\ 
     hidden channels conv.  & 256  & 512  & 128 \\\bottomrule
    \hline
    \end{tabular}
    \caption{Hyperparameter settings of the three different masking models.}
    \label{tab:settings}
\end{table}
\begin{figure}[htb]
\begin{minipage}[b]{1.0\linewidth}
  \centering
  \centerline{\includegraphics[width=4.0cm]{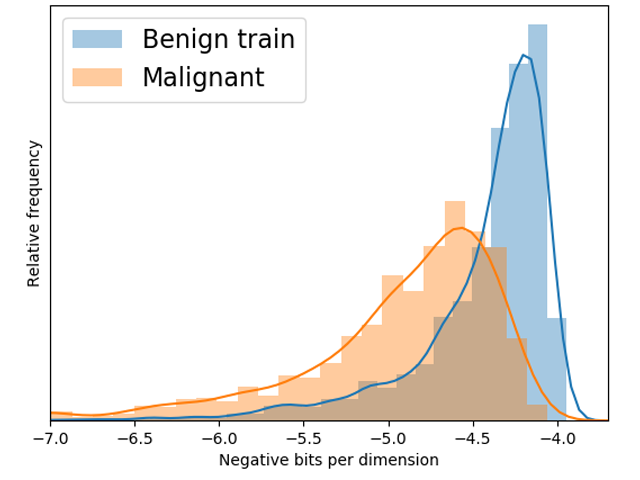}}
  \centerline{\textbf{A} channel-wise}\medskip
\end{minipage}
\begin{minipage}[b]{.48\linewidth}
  \centering
  \centerline{\includegraphics[width=4.0cm]{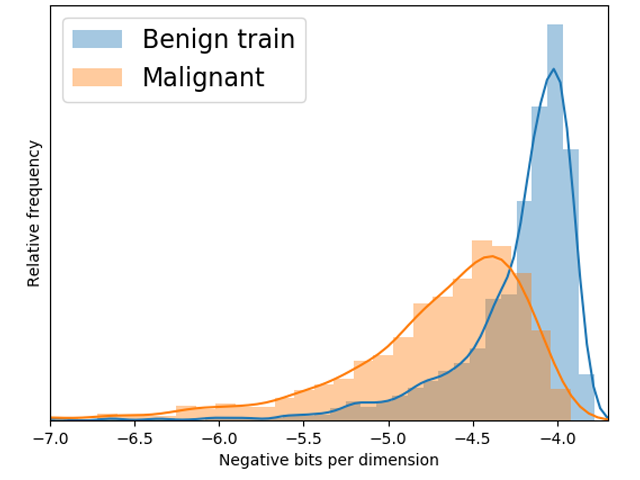}}
  \centerline{\textbf{B} checkerboard}\medskip
\end{minipage}
\hfill
\begin{minipage}[b]{0.48\linewidth}
  \centering
  \centerline{\includegraphics[width=4.0cm]{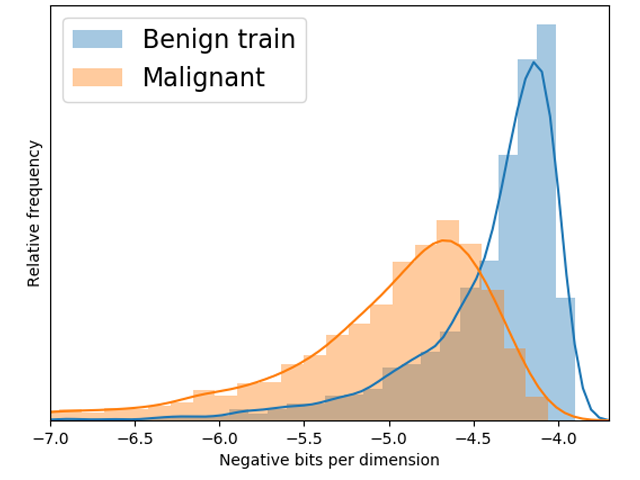}}
  \centerline{\textbf{C} 1-cycle}\medskip
\end{minipage}
\caption{Density estimation of in distribution benign train and out-of-dataset malignant images for multiple masking techniques.}
\label{fig:res}
\end{figure}
\begin{figure}[htb]
\begin{minipage}[b]{.48\linewidth}
  \centering
  \centerline{\includegraphics[width=4.0cm]{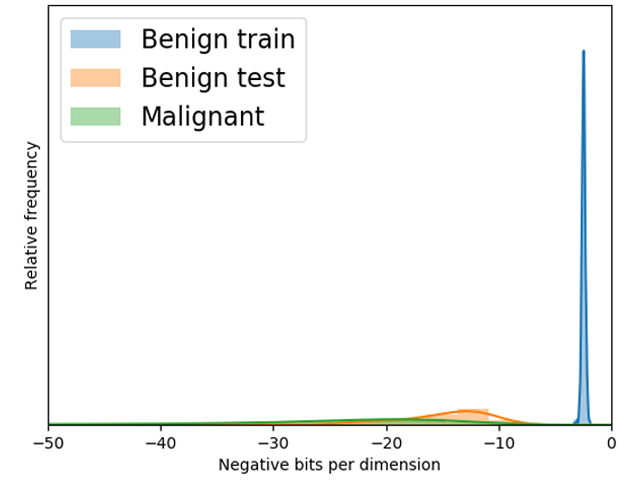}}
  \centerline{\textbf{A} Full view}
\end{minipage}
\hfill
\begin{minipage}[b]{0.48\linewidth}
  \centering
  \centerline{\includegraphics[width=4.0cm]{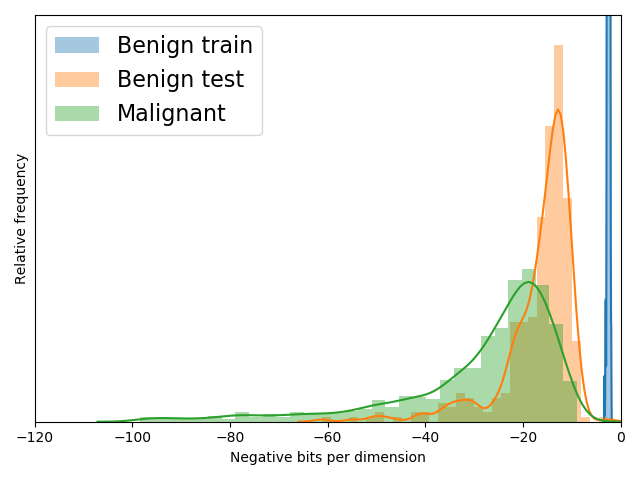}}
  \centerline{\textbf{B} Enhanced}
\end{minipage}
\caption{Likelihood distribution of Wavelet Flow with the ISIC dataset}
\label{fig:ood-wvlt}
\vspace{-0.5cm}
\end{figure}
\begin{figure}[htb]
\begin{minipage}[b]{1.0\linewidth}
  \centering
  \centerline{\includegraphics[width=8.5cm]{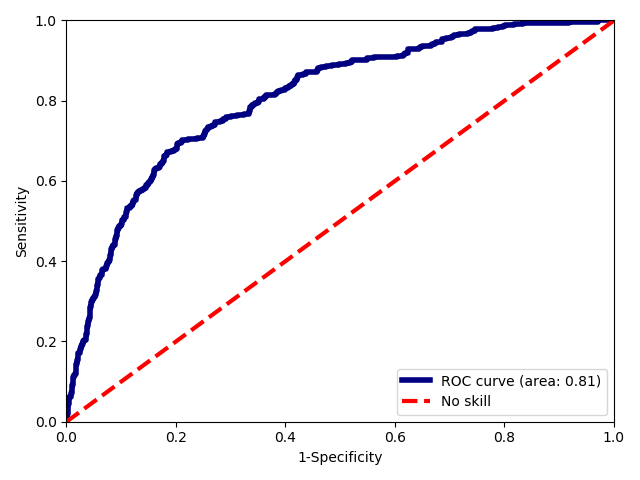}}
  \caption{ROC curve of Wavelet Flow with Benign images from the test set and OOD malignant images.}
  \label{fig:roc2}
\end{minipage}
\end{figure}

We compare in dataset training data with out of dataset data in Figure \ref{fig:res} for the distributions with channel-wise, checkerboard and 1-cycle masking. We find their ROCs to be 0.81, 0.82 and 0.81.

During training and preparation of Wavelet Flow, our computation power and a time constraint limited the number of training iterations for each layer. For full architectural details we suggest reading the original paper. We use the same hyperparameter settings as is given in their \textit{GitHub} repository~\cite{waveletflowgit} except for the fact that we do not randomly crop the images for data augmentation in the final levels and that the training batch size of the final level is 16 instead of 64 due to computational limitations. The number of iterations per layer from 0 to 6 are 1630000, 1130000, 400000, 530000, 510000, 940000 and 840000, respectively. Figure \ref{fig:ood-wvlt} depicts the distribution of the train, test and OOD images. The ROC curve of the test and OOD images is shown in Figure \ref{fig:roc2}. The respective ROC curves of the different models are given in Figure \ref{fig:ood-roc}.
\begin{figure}[htb]
\begin{minipage}[b]{1.0\linewidth}
  \centering
  \centerline{\includegraphics[width=6.5cm]{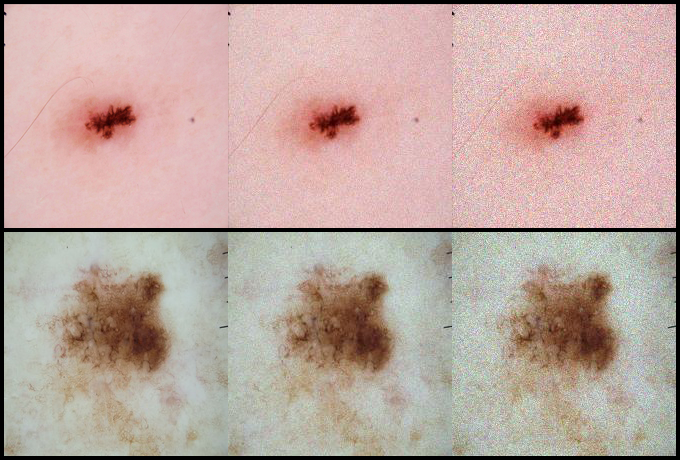}}
  \caption{Benign and malignant images with additive Gaussian noise with variance $0$, $0.05^2$ and $0.1^2$}
  \label{fig:noisy}
\end{minipage}
\end{figure}
\begin{figure}[htb]
\begin{minipage}[b]{1.0\linewidth}
  \centering
  \centerline{\includegraphics[width=7.5cm]{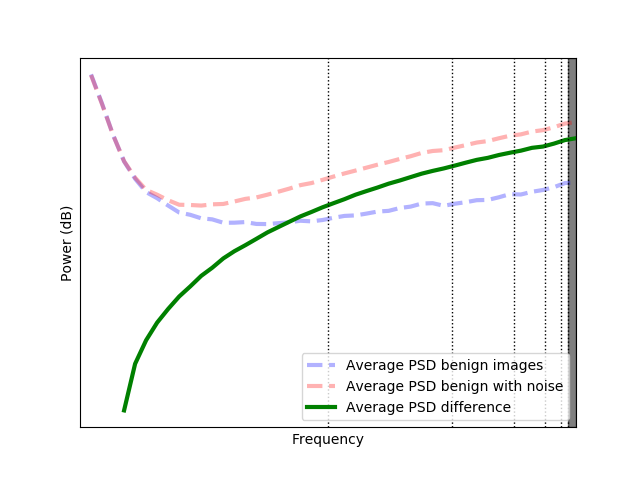}}
  \caption{Average PSD of normal and noisy benign images with multiplicative Gaussian noise of $0.1^2$ variance.}
  \label{fig:psd}
\end{minipage}
\end{figure}
\begin{figure}[htb]
\begin{minipage}[b]{1.0\linewidth}
  \centering
  \centerline{\includegraphics[width=7.5cm]{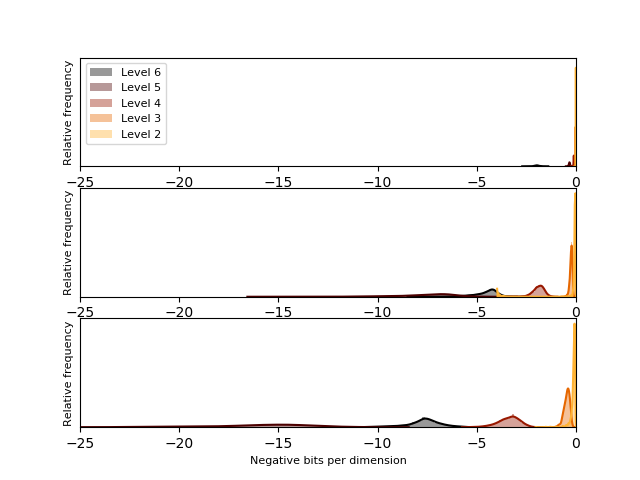}}
  \caption{Per level likelihood estimation of the benign train images with multiplicative Gaussian noise of 0, $0.05^2$ and $0.1^2$ variance from top to bottom. Excluded are level 0 and 1 because their contribution to the total loss is insignificant.}
  \label{fig:ood-fr}
\end{minipage}
\end{figure}
\begin{figure}[htb]
\begin{minipage}[b]{1.0\linewidth}
  \centering
  \centerline{\includegraphics[width=7.5cm]{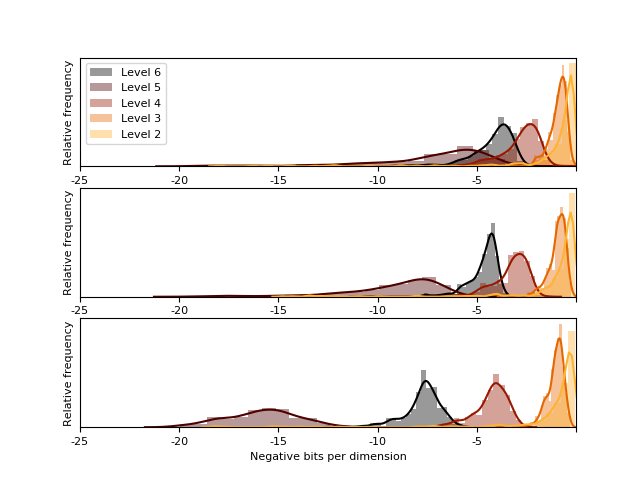}}
  \caption{Per level likelihood estimation of the benign test images with multiplicative Gaussian noise of 0, $0.05^2$ and $0.1^2$ variance from top to bottom. Excluded are level 0 and 1 because their contribution to the total loss is insignificant.}
  \label{fig:ood-fr2}
\end{minipage}
\end{figure}
\begin{figure}[htb]
\begin{minipage}[b]{1.0\linewidth}
  \centering
  \centerline{\includegraphics[width=7.5cm]{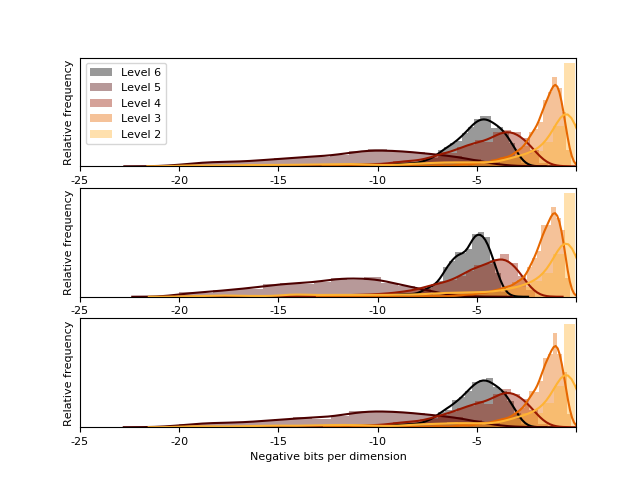}}
  \caption{Per level likelihood estimation of the malignant OOD images with multiplicative Gaussian noise of 0, $0.05^2$ and $0.1^2$ variance from top to bottom. Excluded are level 0 and 1 because their contribution to the total loss is insignificant.}
  \label{fig:ood-fr3}
\end{minipage}
\end{figure}

With the frequency dependent analyses, the first two levels are not considered since its contribution to the BPD is insignificant ($>$0.02) (Appendix \ref{B}). The frequency dependent analysis for the benign test images is given for regular and noisy images. The distribution of images with multiplicative Gaussian noise with variance of $0.05^2$ and $0.1^2$ (see Figure \ref{fig:noisy}) is compared against the original images. We show the same for the malignant OOD data and present the results in Figures \ref{fig:ood-fr}, \ref{fig:ood-fr2} and \ref{fig:ood-fr3}.
We also take the average Power Spectral Density (PSD) (Figure \ref{fig:psd}) of all benign images to identify which level of the Wavelet Flow should theoretically shift the most. The frequency range which the individual levels cover are bounded by vertical lines from left to right.
For completeness, we also compared the generative capabilities of both GLOW and Wavelet Flow (see appendix \ref{C}).

\section{Discussion}
\label{sec:majhead}
We employed GLOW and Wavelet Flow for OOD detection on the ISIC melanoma dataset and find clear differences in performance. As shown in Figure \ref{fig:glowres1}, GLOW generalized well given the large overlap between the train and test distribution. We draw an ROC curve against the malignant distribution that yields an area of 0.82 and 0.81 for the train and test set, respectively. Our initial hypothesis was that the BPD of the train images will be much less than what is actually observed in the results of our experiments as this is found in literature. We also hypothesized that the train set ROC area would have been closer to 1 while the test distribution being less than that (shifted more to the right). This suggests that GLOW generalizes well on the benign set but still finds large similarities with the malignant images and as such, fails to accurately detect them as OOD.  Furthermore, we implemented channel-wise, checkerboard and 1-cycle masking and have shown the likelihood distributions of the benign train and malignant images. This was to find if using masking strategies can lead to better OOD performance. Using the different masking methods the ROC curves of the distributions are 0.81, 0.82 and 0.81, respectively. We do not see any significant change in the ROC curve area by introducing the checkerboard or cycle mask in the coupling layers. The results of our experiments confirm the lack of OOD detection capabilities of GLOW and show that different masking strategies are not the direction to move towards for improvement of it. We believe it is because GLOW is unable to distinguish between semantic and background information as also suggested by previous work. Although these experiments can be improved by having access to better computational power, we do not suggest this approach should be further research as we believe augmentation methods or architectural modifications could potentially lead to more improvement of OOD detection in GLOW.
A modified version of GLOW, Wavelet Flow has recently reported in great improvement over regular GLOW in terms of BPD on the training set. However, from our findings the ROC curve of the test and out-of-dataset data has an area of 0.8 which is similar to GLOW. Furthermore, the training set distribution is very far from the test set distribution. This implies that the model has not generalized well since the train and test images are from the same class. We think enabling random cropping as suggested by the paper can serve as a good regularizer during training. We do not think simply training for more iterations will improve OOD detection. Given that the BPD on the train set is much lower, training Wavelet Flow is much more efficient on high dimensional data and the fact that the ROC curve is similar to GLOW, one can suggest that learning data distributions per frequency component might potentially result in better OOD performance on all datasets compared to GLOW. It might be even more beneficial to consider only particular frequency components since not every layer contributes to the loss equally.  
In the Wavelet Flow implementation, we investigated the likelihood distributions of each wavelet decomposition level independently. As expected, we see that when we add multiplicative Gaussian noise to the benign test images, the higher level curves shift to increasing BPD while lower level curves are almost unaffected. We observe that the likelihood distribution of level 5 shifted the most with increasing noise. The average PSD of the noisy benign images suggests that the sixth level should be subjective to the most shift. This could be the result of the limited time and resource available during the training of the final layer and would then also explain the blurriness of our samples (appendix \ref{C}) since the highest level 6 is responsible for the highest image coefficients. Thus, longer training (potential better convergence) and more consistency between the number of iterations between each layer should translate into more accurate OOD detection of noise and higher quality sampled images. This is in line with literature, since the original paper presents the results with levels trained for up to 5 times more iterations on several datasets. We see for the lower levels (appendix \ref{B}) that only level 1 is affected and as a consequence slightly shifts and flattens while level 0 stays the same in the training set. For the test and out-of-dataset images there is no change to be observed. An additional interesting investigation would be to evaluate the effect of the variance in the higher frequency ranges and the eventual likelihood distributions at each of the individual Wavelet Flow levels. The degree of correlation between these two statistics could potentially serve as a benchmark for frequency dependent likelihood estimation in images.
When the same approach was taken on the malignant data, we found that noise did not impact likelihood estimation of it. Ideally, we would like the likelihood of malignant images to decrease as well with increasing noise. This would then imply that the model has learned the semantics of melanoma very well, understanding that malignant images are not noise or benign but something in between. Our suggestions made for improvement of the combined levels also hold for the frequency dependent OOD detection and encourage for it be explored in future work.
We believe that Wavelet Flow holds promise for OOD detection but needs further improvement. Namely, the test should be closer to the train set distribution. Given the sub-optimal performance of GLOW in our findings, we do not think that the NF used in Wavelet Flow is a big contributor to the improvements made in OOD detection. We see many areas that can be researched to further understand which underlying mechanics enable Wavelet Flow to perform better than GLOW in BPD evaluation and how to improve its OOD detection capability. As mentioned by the authors of Wavelet Flow, Haar wavelets were used but other wavelets should be investigated as well. Also, a different NF architecture, for estimating the detail coefficients, that performs similar or better than GLOW in BPD evaluation can be implemented. If similar or better performance is found, it confirms our hypothesis that the improvements are not due to the choice of NF architecture but, in fact, as a result of the usages of wavelet transformations. If it results in worse results then it means that the decision of choosing a GLOW inspired NF was essential for the performance of the model. Furthermore, using a more state-of-the-art NF architectures might improve OOD detection even further. From literature we know that OOD detection often fails due to the lack of semantic capture and the exploitation of background components of the image. We believe that employing a method that enables the NF model to distinguish between semantic and background features will enable OOD detection using NF the most.
We have found an interesting direction to head for towards tractable likelihood estimation for medical images by implementing frequency based OOD detection for the first time to the best of our knowledge and we mentioned multiple areas for further exploration to gain a better understanding of OOD detection on frequency basis. 

\section{Conclusion}
\label{sec:print}
In this study we apply GLOW and Wavelet Flow for the task of OOD detection on the ISIC dataset. In line with similar findings, we confirm that GLOW lacks ability to reliably detect out-of-distribution data. As further research suggests, we attempt to improve the OOD detection capabilities by employing different masking strategies in the coupling layers of GLOW. These improvements yield little-to-no change in the ability to detect OOD data using GLOW. We further implemented Wavelet Flow which was originally introduced due to the computational efficiency and competitiveness with GLOW in bits per dimension. We have observed a big improvement in BPD evaluation and find similar OOD detection performance. Furthermore, we hypothesize that its ability to capture semantics is better than that of GLOW and therefore see much potential for this model. We have looked at the density and likelihood estimation capabilities of each individual layer in Wavelet Flow and analyzed the effect of random noise on the distributions in order to better understand how well the model generalized. We have done OOD detection on medical images but the applications for this technique is not limited to medical imaging but can be used in other domains as well such as health monitoring or video surveillance. We find NFs in combination with Wavelet decomposition to be very promising given the demand for explainable and interpretable machine learning and suggest further research in to Normalizing Flows with wavelets for density estimation. 


\clearpage 
\printbibliography
\onecolumn
\section*{Appendix}
\appendix
\section{Normalizing Flows derivation}
\label{A}
A bijection between distributions should adhere to the law of total probability as 
    \begin{equation}
        \int p(x)dx= \int p(y)dy.    
    \end{equation}
The absolute change in volume is only relevant. In this fashion we can represent a sample in $Y$ in terms of a sample in $X$ as
\begin{equation}
    p(y)=p(x)\left|\frac{dx}{dy}\right|.
\end{equation}
We now observe two variables in this transformation. We have a Uniform(0,1) distribution being linearly transformed in to any arbitrary parallelogram without shift with the matrix. 
\begin{equation}
    T=\begin{bmatrix}
        a & b \\
        c & d 
    \end{bmatrix}
\end{equation}
Now, the change in absolute volume of the parallelogram is
\begin{equation}
    \left|ad-cb\right|=\left|\operatorname{det}(T)\right|.
\end{equation}
Now for a multidimensional non-linear transformation, one approximates this with infinite, infinitesimally parallelograms to track the global change in volume. This local-linear change approximation is in fact the Jacobian determinant. Given that $y=f(x)$ thus $x=f^{-1}(y)$ we can write
\begin{equation}
    p_Y(y)=p_X(f^{-1}(y))\left|\operatorname{det}\frac{\partial f^{-1}}{\partial x}\right|.
\end{equation}
Based on a chain of these operations as has been shown in Figure \ref{fig:flow} one can write
\begin{subequations}
    \begin{align}
    \mathbf{z}_{i-1} & \sim p_{i-1}\left(\mathbf{z}_{i-1}\right) \\
    \mathbf{z}_{i} &=f_{i}\left(\mathbf{z}_{i-1}\right),  \mathbf{z}_{i-1}=f_{i}^{-1}\left(\mathbf{z}_{i}\right)
    \end{align}
\end{subequations}
and thus estimate the probability as
\begin{equation}
\begin{aligned}
    p_{i}\left(\mathbf{z}_{i}\right) &=p_{i-1}\left(f_{i}^{-1}\left(\mathbf{z}_{i}\right)\right)\left|\operatorname{det} \frac{d f_{i}^{-1}}{d \mathbf{z}_{i}}\right|. 
\end{aligned} 
\end{equation}
The inverse function theorem and the property $\operatorname{det}(\matr{M}^{-1})=\operatorname{det}(\matr{M})^{-1}$ for invertible matrices $\matr{M}$ allows us to rewrite the transformation as
\begin{subequations}
\begin{align}
    p_{i}\left(\mathbf{z}_{i}\right) &=p_{i-1}\left(f_{i}^{-1}\left(\mathbf{z}_{i}\right)\right)\left|\operatorname{det} \frac{d f_{i}^{-1}}{d \mathbf{z}_{i}}\right| \\
    &=p_{i-1}\left(\mathbf{z}_{i-1}\right)\left|\operatorname{det}\left(\frac{d f_{i}}{d \mathbf{z}_{i-1}}\right)^{-1}\right| \\
    &=p_{i-1}\left(\mathbf{z}_{i-1}\right)\left|\operatorname{det} \frac{d f_{i}}{d \mathbf{z}_{i-1}}\right|^{-1}
\end{align}
\end{subequations}
and we can state the likelihood of a transformation as
\begin{equation}
    \begin{aligned}
\log p_{i}\left(\mathbf{z}_{i}\right) &=\log p_{i-1}\left(\mathbf{z}_{i-1}\right)-\log \left|\operatorname{det} \frac{d f_{i}}{d \mathbf{z}_{i-1}}\right|.
\end{aligned}
\end{equation}
Given the chain of the transformations
\begin{equation}
\mathbf{x}=\mathbf{z}_{K} =f_{K} \circ f_{K-1} \circ \cdots f_{1}\left(\mathbf{z}_{0}\right) \\
\end{equation}
and the log-likelihood of the observed data $\vect{x}$
\begin{subequations}
\begin{align}
\log p(\mathbf{x})=\log p_{K}\left(\mathbf{z}_{K}\right) &=\log p_{K-1}\left(\mathbf{z}_{K-1}\right)-\log \left|\operatorname{det} \frac{d f_{K}}{d \mathbf{z}_{K-1}}\right| \\
&=\log p_{K-2}\left(\mathbf{z}_{K-2}\right)-\log \left|\operatorname{det} \frac{d f_{K-1}}{d \mathbf{z}_{K-2}}\right|-\log \left|\operatorname{det} \frac{d f_{K}}{d \mathbf{z}_{K-1}}\right| \\
&=\ldots \notag
\end{align}
\end{subequations}
or concisely written as
\begin{equation}
    \begin{aligned}
        \log p(\mathbf{x})=\log p_{0}\left(\mathbf{z}_{0}\right)-\sum_{i=1}^{K} \log \left|\operatorname{det} \frac{d f_{i}}{d \mathbf{z}_{i-1}}\right|
    \end{aligned}
\end{equation}

\clearpage 
\section{Low level frequency dependent distributions}
\label{B}
\begin{figure}[htp]
\begin{minipage}[b]{1.0\linewidth}
  \centering
  \centerline{\includegraphics[width=9cm]{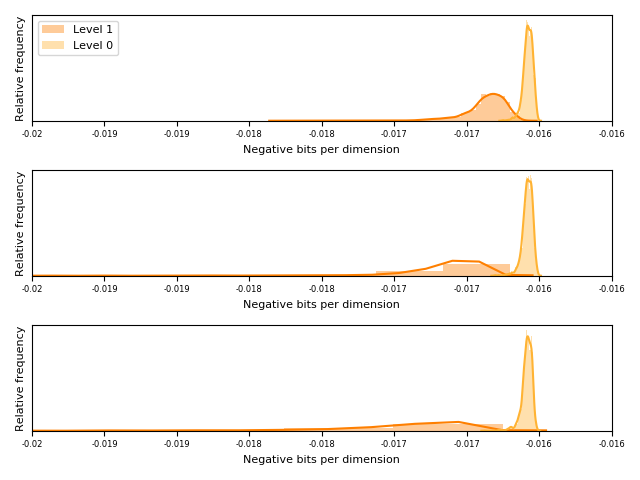}}
  \caption{Low level likelihood estimations of the malignant out-of-dataset images with multiplicative Gaussian noise of 0, $0.05^2$ and $0.1^2$ variance from top to bottom.}
  \label{fig:ood-fr4}
\end{minipage}
\end{figure}
\begin{figure}[htp]
\begin{minipage}[b]{1.0\linewidth}
  \centering
  \centerline{\includegraphics[width=9cm]{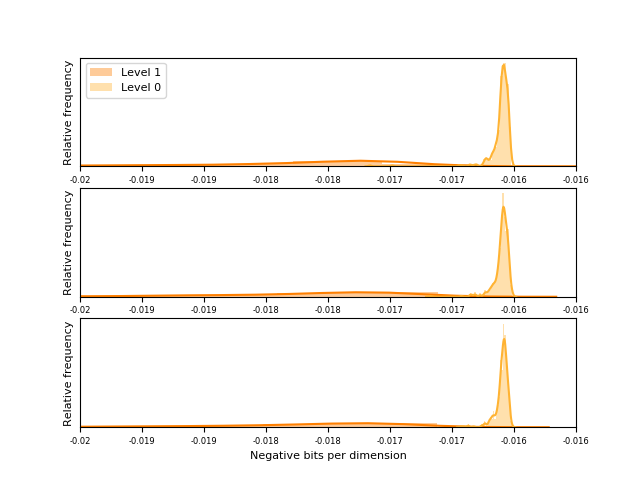}}
  \caption{Low level likelihood estimations of the malignant out-of-dataset images with multiplicative Gaussian noise of 0, $0.05^2$ and $0.1^2$ variance from top to bottom.}
  \label{fig:ood-fr5}
\end{minipage}
\end{figure}
\begin{figure}[htp]
\begin{minipage}[b]{1.0\linewidth}
  \centering
  \centerline{\includegraphics[width=9cm]{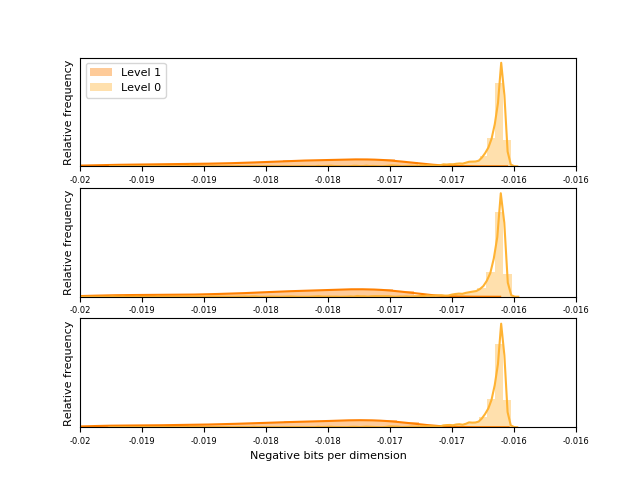}}
  \caption{Low level likelihood estimations of the malignant out-of-dataset images with multiplicative Gaussian noise of 0, $0.05^2$ and $0.1^2$ variance from top to bottom.}
  \label{fig:ood-fr6}
\end{minipage}
\end{figure}

\clearpage 
\section{Samples from GLOW and Wavelet Flow}
\label{C}

\begin{figure}[hbp]
\begin{minipage}[b]{.48\linewidth}
  \centering
  \centerline{\includegraphics[width=7.0cm]{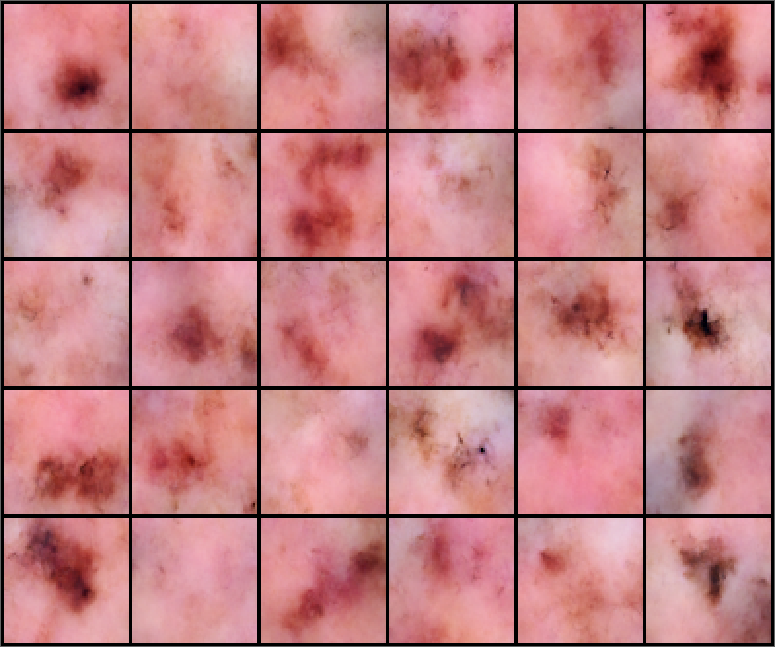}}
  \centerline{\textbf{A} GLOW}
\end{minipage}
\hfill
\begin{minipage}[b]{0.48\linewidth}
  \centering
  \centerline{\includegraphics[width=7.0cm]{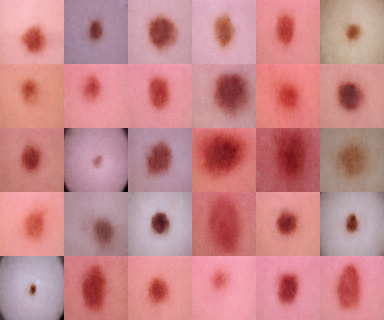}}
  \centerline{\textbf{B} Wavelet Flow}
\end{minipage}
\caption{Generated images representing benign melanoma}
\label{fig:ood-wvlt2}
\vspace{-0.5cm}
\end{figure}

\end{document}